\documentclass[10pt,twocolumn,letterpaper]{article}

\usepackage{cvpr}              

\usepackage{graphicx}
\usepackage{amsmath}
\usepackage{amssymb}
\usepackage{booktabs}

\usepackage{multirow}
\usepackage{caption}
\usepackage{subcaption}
\usepackage{array}
\usepackage{pbox}

\usepackage[pagebackref,breaklinks,colorlinks]{hyperref}

\usepackage[capitalize]{cleveref}
\crefname{section}{Sec.}{Secs.}
\Crefname{section}{Section}{Sections}
\Crefname{table}{Table}{Tables}
\crefname{table}{Tab.}{Tabs.}

\begin{document}

\title{Adversarial Examples on Segmentation Models Can be Easy to Transfer}

\author{Jindong Gu$^1\;\;\;\;\;\;\;\;$  Hengshuang Zhao$^2\;\;\;\;\;\;\;\;$  Volker Tresp$^1\;\;\;\;\;\;\;\;$  Philip Torr$^2$\\
\\
$^1$University of Munich$\;\;\;$  $^2$Torr Vision Group, University of Oxford\\
}

\maketitle

\begin{abstract}
Deep neural network-based image classification can be misled by adversarial examples with small and quasi-imperceptible perturbations. Furthermore, the adversarial examples created on one classification model can also fool another different model. The transferability of the adversarial examples has recently attracted a growing interest since it makes black-box attacks on classification models feasible. As an extension of classification, semantic segmentation has also received much attention towards its adversarial robustness. However, the transferability of adversarial examples on segmentation models has not been systematically studied. In this work, we intensively study this topic. First, we explore the overfitting phenomenon of adversarial examples on classification and segmentation models. In contrast to the observation made on classification models that the transferability is limited by overfitting to the source model, we find that the adversarial examples on segmentations do not always overfit the source models. Even when no overfitting is presented, the transferability of adversarial examples is limited. We attribute the limitation to the architectural traits of segmentation models, i.e., multi-scale object recognition. Then, we propose a simple and effective method, dubbed dynamic scaling, to overcome the limitation. The high transferability achieved by our method shows that, in contrast to the observations in previous work, adversarial examples on a segmentation model can be easy to transfer to other segmentation models. Our analysis and proposals are supported by extensive experiments.
\end{abstract}

\section{Introduction}
Small and quasi-imperceptible input perturbations can mislead the predictions of deep neural networks~\cite{szegedy2013intriguing,goodfellow2014explaining}. Input examples with such artificial perturbations are called adversarial examples (AEs). The standard deep neural networks based image classification models are all vulnerable to the AEs, e.g. VGG~\cite{simonyan2014very} and ResNet~\cite{he2016deep}. In past years, the adversarial examples on image classifications have been intensively studied from various perspectives. Especially, the AEs created on a classification model can also fool other classification models. Recently, the transferability of AEs has raised research interest since it makes practical black-box attacks feasible.

However, the transferability of AEs can be limited by overfitting to the employed source classification model. Concretely, the AEs are often created by the multi-step attack method~\cite{kurakin2016adversarial}. The AEs will underfit source models when too few attack iterations are applied. Contrarily, a large number of attack iterations will lead to overfitting to the source model. In order to improve the transferability of AEs on classification models, many efforts have been made to overcome the overfitting problem, i.e.~\cite{dong2018boosting,xie2019improving,dong2019evading,huang2019enhancing,lin2019nesterov,inkawhich2020perturbing,wang2020unified,zou2020improving,guo2020backpropagating,wu2020skip,li2020learning,wu2021improving}, to name a few.

As an extension to image classification task, image semantic segmentation aims to classify all individual pixels of input images. The segmentation models are often deployed in many real-world safety-critical applications, e.g. medical domain~\cite{milletari2016v} and autonomous driving systems~\cite{kaymak2019brief}. Thus, the adversarial robustness of segmentation models has also received much attention recently~\cite{xie2017adversarial,arnab2018robustness,xu2021dynamic,xiao2018characterizing,hendrik2017universal,he2019non,nesti2021evaluating,fischer2021scalable,paschali2018generalizability}. It has been observed in previous work that the AEs on segmentation models are difficult to transfer across different network architectures and scales. However, a comprehensive investigation into the transferability of AEs on segmentation models has not been done yet. Therefore, we focus on this topic in this work.

The feature extractors of the DNN-based classification framework are often taken as backbones to build segmentation models. For comparison, we first investigate the transferability of AEs on the classification models that correspond to segmentation backbones. Our investigation shows that the AEs on all popular classification models will overfit the source model when a large number of attack iterations are applied, e.g., more than 20 iterations. The degree of overfitting depends on the model architectures. Specifically, the AEs on ResNet~\cite{he2016deep} with skip connections show more drastic overfitting than the ones on VGG~\cite{simonyan2014very}.

Next, we explore the overfitting phenomenon of AEs on segmentation models. We find that the observation made on image classifications does not hold on image segmentations anymore. Namely, the AEs do not always overfit the source segmentation model. Specifically, the degree of overfitting depends on the backbone used by the segmentation models. When ResNet backbone is applied in the segmentation models, the AEs can quickly overfit the source model. In contrast, in the case of a segmentation model with VGG backbone, the AEs do not overfit at all, even after 10k attack iterations. The findings is interesting since the number of parameters of the VGG backbone is significantly larger than in the ResNet backbone.

Even when no overfitting is presented, the transferability of AEs is still limited. We attribute the further limitation to the architectural traits of segmentation models. Segmentation models often apply multiple scales to recognize objects of different sizes, e.g., with dilated convolution~\cite{chen2017deeplab,yu2015multi} or specialized pooling~\cite{chen2017deeplab,zhao2017pyramid}. Based on the architectural traits of segmentation models as well as our explorations, we propose a simple and effective method, called the dynamic scaling. Concretely, we apply a large number of attack iterations and randomly scale the input images and ground-truth masks in each attack iteration. By doing this, we can create AEs that transfer well to different scales of different segmentation models. In contrast to observations made in previous work~\cite{xie2017adversarial,arnab2018robustness}, the high transferability achieved by our method shows that adversarial examples on a segmentation model can be easy to transfer to others.

Our contributions can be summarized as follows:
\begin{itemize}
   \item We first investigate the transferability of AEs on classification models that correspond to the popular segmentation backbones. Our investigation shows the AEs on all involved classification models will overfit the source models and reveal the relationship between the degree of overfitting and model architectures.
  \item We then explore the overfitting phenomenon of AEs on segmentation models. We find that the AEs do not always overfit source segmentation models. The AEs on segmentation models with VGG backbone do not overfit at all, while the ones on ResNet backbone-based segmentation models overfit drastically and quickly.
  \item Based on our investigation and architectural traits of segmentation models, we propose a simple and effective method to improve the transferability of AEs on segmentation models. This work is the first to make AEs transferable between segmentation models.
\end{itemize}

\section{Related Work}
\paragraph{Adversarial robustness on segmentation models.}
The work~\cite{arnab2018robustness} makes an extensive study on the adversarial robustness of segmentation models and demonstrates the inherent robustness therein. Especially, they find that adversarial examples do not transfer well across different scales and transformations. The work~\cite{xie2017adversarial} extends adversarial examples to semantic segmentation and object detection. They also show that the adversarial examples created by their attack method do not transfer well across different network structures. Besides, the adversarial robustness of segmentation models has also been well studied from other perspectives, such as adversarial training~\cite{xu2021dynamic}, adversarial example detection~\cite{xiao2018characterizing}, and universal adversarial perturbation~\cite{hendrik2017universal}. The adversarial robustness of segmentation has received much attention recently. In previous experiments~\cite{xie2017adversarial,arnab2018robustness}, it is observed that the AEs on segmentation models are difficult to transfer. The experiments show segmentation models are robust to defend transfer-based black-box attacks. However, a comprehensive exploration of the transferability of AEs has not been done yet in semantic segmentation.

\vspace{0.2cm}\noindent\textbf{Transferability of adversarial examples on classification models. }
The transfer of AEs between classification models has been intensively investigated. When too many attack iterations are applied, the created adversarial examples are overfitted to the source classification model. To achieve high transferability, many techniques have been proposed to mitigate the overfitting phenomenon from the perspective of data, models, loss design, and optimization methods. Concretely, the work ~\cite{xie2019improving,dong2019evading,zou2020improving,wu2021improving} creates AEs by optimizing the loss on a single or a set of transformed images. Besides,~\cite{guo2020backpropagating,wu2020skip,li2020learning} modify the optimization process according to the model architecture. To regularize the optimization process, new losses formulated in feature space are also proposed~\cite{huang2019enhancing,inkawhich2020perturbing,wang2020unified}. Furthermore, more advanced optimization algorithms, like the momentum method~\cite{dong2018boosting,polyak1964some} and the Nesterov Accelerated Gradient~\cite{lin2019nesterov,Nesterov1983AMF} have also been explored to prevent overfitting of AEs. Even though many efforts have been made to overcome the overfitting problem, little attention has been paid to the number of attack iterations. Due to the overfitting phenomenon, it is difficult to set the number of attack iterations since the target classification models are assumed to be unavailable.

\section{Adversarial Attack Methods}
Adversarial examples for a given model can be generated by leveraging the gradient information of the inputs. The single-step FGSM~\cite{goodfellow2014explaining} attack method creates adversarial examples by taking a one-step update. Furthermore, the multi-step adversarial attack would cause a more serious threat, e.g., BIM~\cite{kurakin2016adversarial}. Hence, the multi-step adversarial attack is more often applied to create AEs. In this section, we introduce the application of BIM attack to classification and segmentation models.

\vspace{-0.2cm}\paragraph{Adversarial attack on image classification.}
In image classification, given the classification model $f_{cls}(\cdot)$, the clean image $\boldsymbol{X}^{clean}\in\mathbb{R}^{{H}\times{W}\times{C}}$ and its one-hot label $\boldsymbol{y}\in\mathbb{R}^{1\times{N}}$, the image is classified as $f_{cls}(\boldsymbol{X}^{clean})\in\mathbb{R}^{1\times{N}}$. The goal of attack is to create an adversarial example $\boldsymbol{X}^{adv}$ to mislead the model, i.e., $argmax(f_{cls}(\boldsymbol{X}^{adv}))\neq argmax(\boldsymbol{y})$ where $argmax(\cdot)$ outputs the index of the maximum. BIM creates AEs on the classification model via multiple iterations in Equation~\ref{equ:bim}.

{\footnotesize
\begin{equation}
\boldsymbol{X}^{adv_{t+1}} = \phi^{\epsilon}(\boldsymbol{X}^{adv_{t}} + \alpha\cdot\textit{sign}(\nabla_{\boldsymbol{X}^{adv_{t}}} L(f_{cls}(\boldsymbol{X}^{adv_{t}}), \boldsymbol{y}))),
\label{equ:bim}
\end{equation}
}where $\alpha, \epsilon$ are the step size and the perturbation range, respectively. $\boldsymbol{X}^{adv_t}$ is the AE after the $t$-th attack step, and the initial value is set to $\boldsymbol{X}^{adv_0} = \boldsymbol{X}^{clean}$. The $\phi^{\epsilon}(\cdot)$ function clips its output into the range $[\boldsymbol{X}^{clean}-\epsilon, \boldsymbol{X}^{clean}+\epsilon]$. $sign(\cdot)$ is the sign function and $\nabla_a(b)$ is the matrix derivative of b with respect to a. $L(\cdot)$ stands for the cross-entropy loss function,
{\footnotesize
\begin{equation}
L(f_{cls}(\boldsymbol{X}^{adv_{t}}), \boldsymbol{y}) = CE(f_{cls}(\boldsymbol{X}^{adv_{t}}), \boldsymbol{Y}).
\label{equ:cls_loss}
\end{equation}
}The AEs created with more attack iterations often lead to a higher attack success rate on the source model. However, when transferring the created AEs to target classification models, the number of iterations is often empirically and carefully selected.

\paragraph{Adversarial attack on semantic segmentation.} In semantic segmentation, given the segmentation model $f_{seg}(\cdot)$, the clean image $\boldsymbol{X}^{clean}\in\mathbb{R}^{{H}\times{W}\times{C}}$ and its segmentation label $\boldsymbol{Y}\in\mathbb{R}^{{H}\times{W}\times{N}}$, the segmentation model classifies all individual pixels of the input image $f_{seg}(\boldsymbol{X}^{clean})\in\mathbb{R}^{{H}\times{W}\times{N}}$. The goal of attack is to create the adversarial example $\boldsymbol{X}^{adv}$ to mislead all the pixel classifications, i.e., $argmax(f_{seg}(\boldsymbol{X}^{adv})_i)\neq argmax(\boldsymbol{Y}_i)$ where $i\in [1, {H}\times{W}]$ corresponds to the index of a input pixel. BIM attack can also be easily applied to create an AE on semantic segmentation. Similar to Equation~\ref{equ:bim}, AEs can be created with the following Equation~\ref{equ:bim_seg}.

{\footnotesize
\begin{equation}
\boldsymbol{X}^{adv_{t+1}} = \phi^{\epsilon}(\boldsymbol{X}^{adv_{t}} + \alpha\cdot\textit{sign}(\nabla_{\boldsymbol{X}^{adv_{t}}} L(f_{seg}(\boldsymbol{X}^{adv_{t}}), \boldsymbol{Y})))
\label{equ:bim_seg}
\end{equation}
}The difference of this Equation from the one on classification is the loss term. Concretely, the loss term in the segmentation is as follows
{\footnotesize
\begin{equation} \hspace{-0.05cm}
L(f_{seg}(\boldsymbol{X}^{adv_{t}}), \boldsymbol{Y}) = \frac{1}{{H}\times{W}} \sum_{i=1}^{{H}\times{W}} CE(f_{seg}(\boldsymbol{X}^{adv_{t}})_i, \boldsymbol{Y}_i),
\label{equ:seg_loss}
\end{equation}
}where $(H, W)$ is the input size. The loss is obtained by averaging the cross-entropy losses over all individual pixels of the input image.

\section{Transferability of AEs on Classification}
For comparison, we first investigate the transferability of AEs on classifications. Especially, we take VGG16~\cite{simonyan2014very} and ResNet50~\cite{he2016deep} as source models, respectively. The feature extractors of the two classification models are often used as backbones in segmentation models. We test other popular classification models on the AEs created on the source model to evaluate the transferability of AEs.

\paragraph{Experimental setting.} The popular classification models are included in this experiment, namely, VGG~\cite{simonyan2014very}, ResNet~\cite{he2016deep}, GoogleNet~\cite{GoogleNetSzegedyLJSRAEVR14}, MobileNet~\cite{howard2017mobilenets} and MNasNet~\cite{tan2019mnasnet}. The pre-trained classification models are taken from PyTorch library~\cite{NEURIPS2019_9015}. The ImageNet 1k dataset~\cite{deng2009imagenet} is used. We randomly select 10k images from the validation dataset for evaluation. The standard protocol to create AEs is followed. Given the image range [0, 1], the perturbation range $\epsilon$ is set to 0.03 and the step size $\alpha$ is 0.01. Since the standard mIoU is applied to evaluate the transferability of AEs on segmentation, we take the standard classification accuracy (in \%) to evaluate the transferability of AEs. The high performance of a target classifier on AEs means that the low transferability of the AEs.

\begin{figure*}[!h]
    \begin{subfigure}[b]{0.48\textwidth}
    \centering
    \includegraphics[scale=0.68]{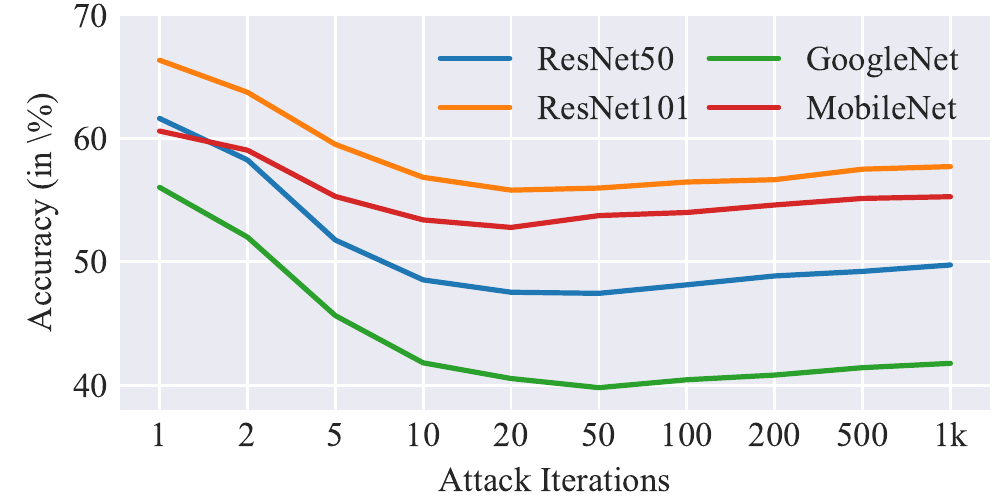}
    \caption{\footnotesize VGG16 as Source Model}
    \label{subfig:vgg_overfit}
    \end{subfigure}  \hspace{1.3em}
    \begin{subfigure}[b]{0.48\textwidth}
    \centering
    \includegraphics[scale=0.68]{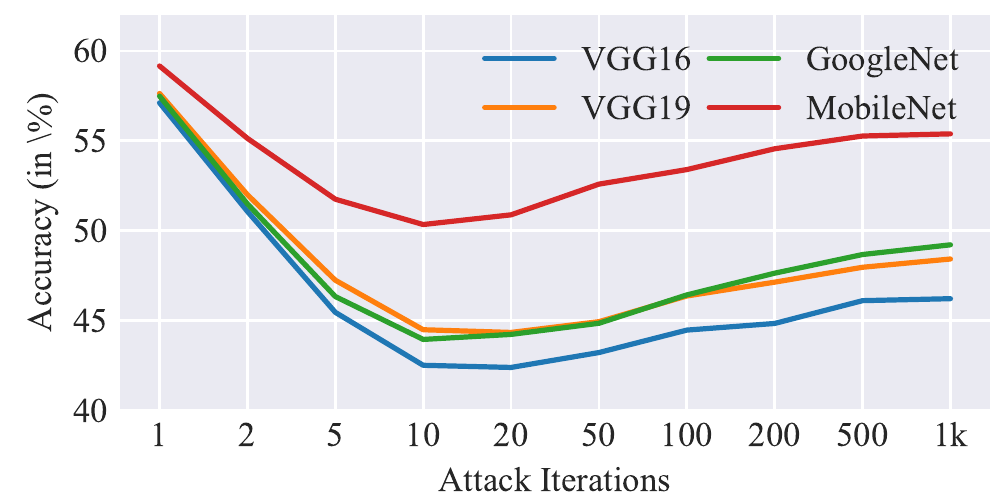}
    \caption{\footnotesize ResNet50 as Source Model}
    \end{subfigure}
    \begin{subfigure}[b]{0.48\textwidth}
    \centering
    \includegraphics[scale=0.68]{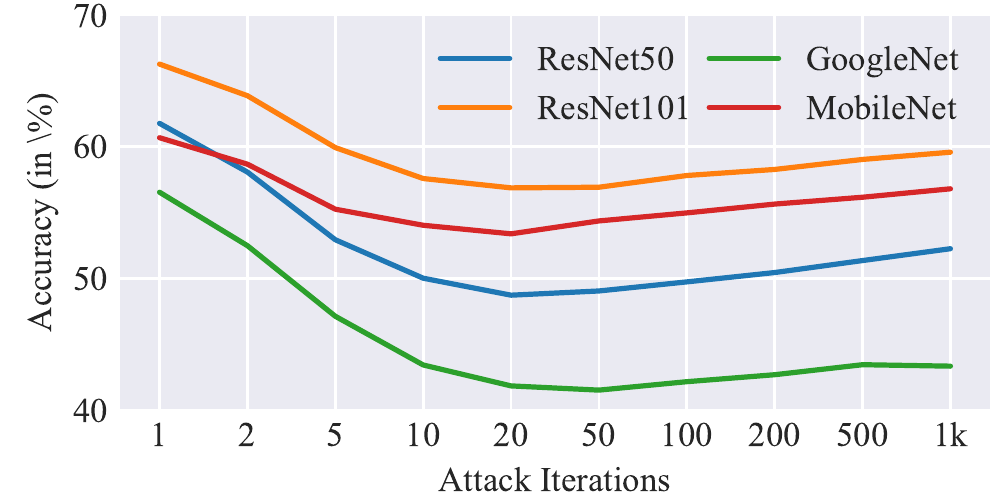}
    \caption{\footnotesize VGG19 as Source Model}
    \end{subfigure}  \hspace{1.5em}
    \begin{subfigure}[b]{0.48\textwidth}
    \centering
    \includegraphics[scale=0.68]{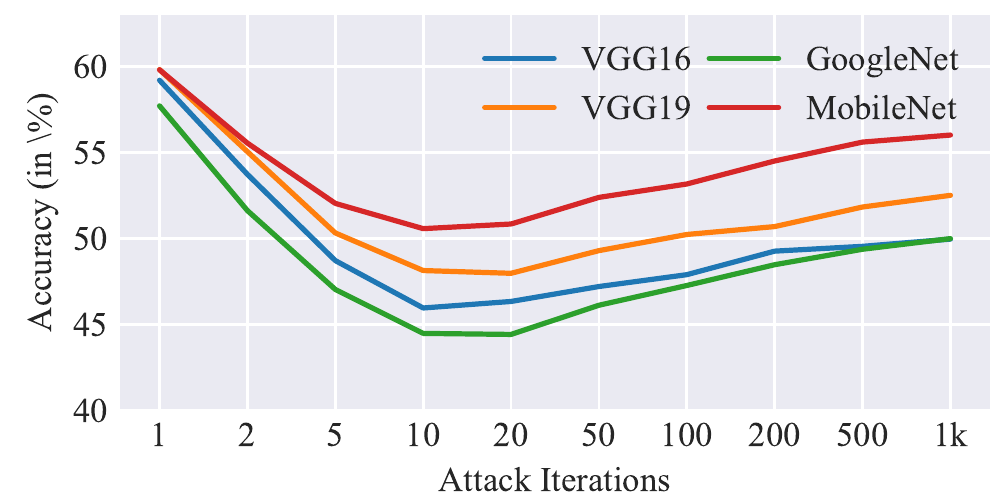}
    \caption{\footnotesize ResNet101 as Source Model}
    \end{subfigure}
    \caption{Transferability of adversarial examples on classification models. The high accuracy of target models on AEs means the poor transferability of the AEs. In all subfigures, the transferability of the AEs becomes better from the first after a few iterations. When a large number of attack iterations (e.g. over 100) are applied, the transferability of AEs becomes poor again. The overfitting phenomenon on ResNet in subfigures (b) and (d) is more drastic than that on VGG in subfigures (a) and (c).}
    \label{fig:cls_overfit}
\end{figure*}

\paragraph{Experimental results and analysis.} The results are reported in Fig.~\ref{fig:cls_overfit}. In each subfigure, we show the classification accuracy of target classifiers on adversarial examples in given attack iterations. For instance, in Fig.~\ref{subfig:vgg_overfit}, the VGG16 is taken as source model, ResNet50, ResNet101, GoogleNet and MobileNet are tested as target models. Note that we only show the performance on the target models since the performance on the source model is close to zero.

When only a few attack iterations (e.g. 2) are applied to create AEs, the created AEs underfit the parameters of source models. As shown in all subfigures of Fig.~\ref{fig:cls_overfit}, the performance on the target classifiers is high, which indicates that the transferability of AEs is low. Contrarily, when too many attack iterations (e.g. 100) are applied, the AEs overfit the parameters of source models, which also leads to lower transferability. The overfitting phenomenon can be observed on both VGG and ResNet models.

\begin{figure}[t]    
    \centering
    \includegraphics[scale=0.7]{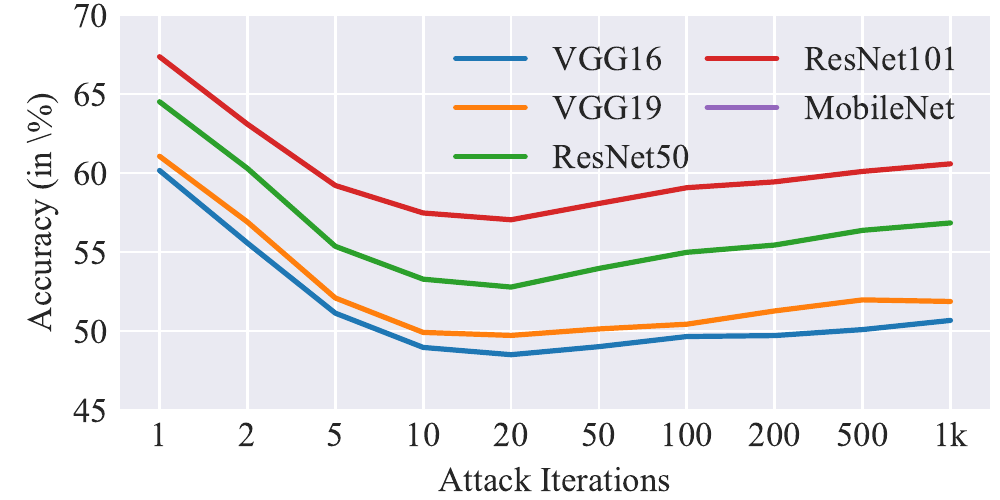}
    \caption{Transferability of adversarial on GoogleNet. The degree of overfitting on this model is somewhere between VGG and ResNet. See Fig.~\ref{fig:cls_overfit} for a comparison.} \vspace{-0.1cm}
    \label{fig:google_overfit}
\end{figure}

Besides, we find that the AEs on ResNet overfit more drastically than the ones on VGG models (similar observations on VGG19 and ResNet101). The drastic overfitting on ResNet can be attributed to the residual block. Instead of the skip connection, the AEs on ResNet start to overfit the residual connections when more attack steps are applied. The mapping learned by residual connections is different from the ones learned by other networks. Hence, AEs on ResNet overfit drastically. Our arguments are also supported by the work~\cite{wu2020skip}. In GoogleNet or MobileNet, there is no skip connection. They are equipped with multiple paths in each building block. The degree of overfitting on these models is less drastic than ResNet, but slightly more drastic than VGG. See Fig.~\ref{fig:google_overfit} for an example.

As shown in Fig.~\ref{fig:cls_overfit}, the transferability of AEs on classification is sensitive to the number of attack iterations. In current literature, there is no clear guidance on how to set the number of attack iterations to achieve the high transferability of AEs. The number of attack iteration is often empirically selected, e.g., 10 or 20~\cite{xie2019improving,dong2019evading,zou2020improving}. As shown in Fig.~\ref{fig:cls_overfit}, the empirically selected number of attack iteration happens to be optimal. However, different from classification, semantic segmentation classifies all input pixels instead of a single object in the input image. It is not clear how sensitive the transferability of AEs on segmentation models is to the number of attack iterations. Furthermore, it remains open whether a similar overfitting phenomenon can be observed on the AEs on segmentation models.

\begin{figure*}[t]    
    \begin{subfigure}[b]{0.48\textwidth}
    \centering
    \includegraphics[scale=0.68]{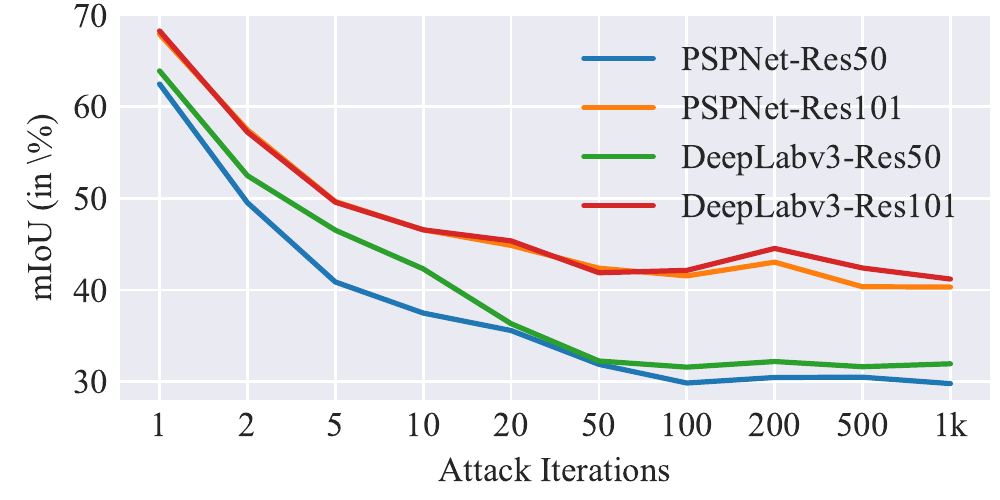}
    \caption{\footnotesize FCN8s-VGG16 as Source Model}
    \label{fig:seg_fcn_overfit}
    \end{subfigure}  \hspace{1.3em}
    \begin{subfigure}[b]{0.48\textwidth}
    \centering
    \includegraphics[scale=0.68]{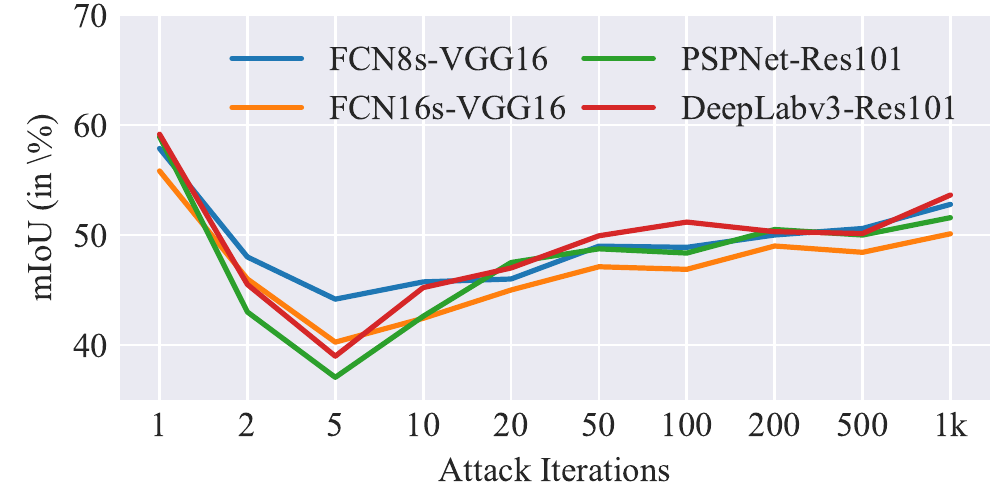}
    \caption{\footnotesize PSPNet-Res50 as Source Model}
    \end{subfigure}
    \begin{subfigure}[b]{0.48\textwidth}
    \centering
    \includegraphics[scale=0.68]{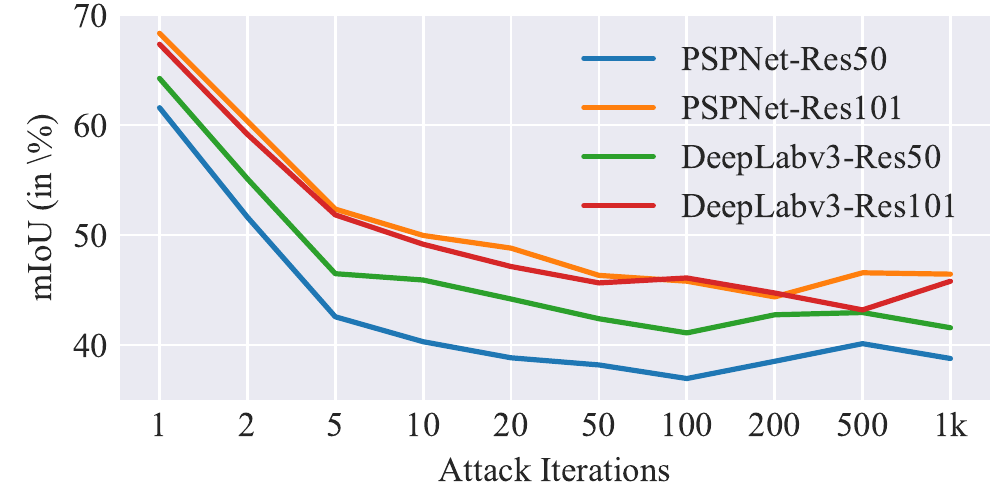}
    \caption{\footnotesize FCN16s-VGG16 as Source Model}
    \label{fig:seg_fcn16s_overfit}
    \end{subfigure}  \hspace{1.3em}
    \begin{subfigure}[b]{0.48\textwidth}
    \centering
    \includegraphics[scale=0.68]{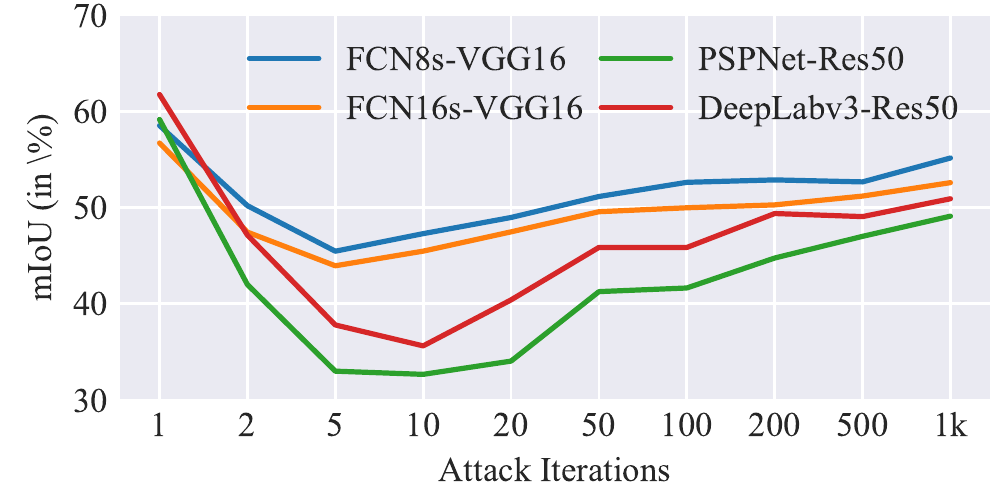}
    \caption{\footnotesize PSPNet-Res101 as Source Model}
    \label{fig:seg_psp101_overfit}
    \end{subfigure}
    \caption{Transferability of adversarial examples on segmentation models. The adversarial examples on segmentation models with ResNet backbones quickly overfit the source segmentation models. In contrast, no overfitting phenomenon is observed on segmentation models with VGG backbones in subfigures (a) and (c).}
    \label{fig:seg_overfit}
\end{figure*}

\section{Transferability of AEs on Segmentation}
In the last section, we show the overfitting of AEs on classification and reveal the relationship between the degree of overfitting and model architectures. Especially, we point out the dilemma in the selection of the number of attack iterations. In this section, we investigate the transferability of AEs on segmentation from similar perspectives.

\paragraph{Experimental setting.} Three semantic segmentation architectures are taken as base models, namely, the popular architectures PSPNet~\cite{zhao2017pyramid} and DeepLabv3~\cite{chen2017rethinking} built on ResNet backbones, and the primary semantic segmentation architecture FCN~\cite{long2015fully} built on VGG backbone. We denote the PSPNet built on ResNet50 backbone as PSPNet-Res50. The PASCAL VOC 2012 dataset (with abbreviation as VOC)~\cite{everingham2010pascal} and Cityscapes dataset~\cite{cordts2016cityscapes} are used in this work. The training data therein are used to train segmentation models in a standard way, respectively. The AEs are created for the clean images in the validation datasets. Similar to the protocol used in the classification, given the image range [0, 1], the perturbation range $\epsilon$ is set to 0.03 for all pixels and the step size $\alpha$ is 0.01.  The standard segmentation evaluation metric mIoU (in \%) is used to evaluate the performance of segmentation models. The low mIoU of target segmentation models on AEs created on the source model means high transferability of the AEs. 

\vspace{-0.2cm}\paragraph{Experimental results and analysis.} The mIoU of target segmentation models on adversarial examples is reported in Fig.~\ref{fig:seg_overfit}. In each subfigure, we show the mIoU of target segmentation models on AEs in given attack iterations. For instance, in Fig.~\ref{fig:seg_fcn_overfit}, the FCN8s-VGG16 is taken as source model, PSPNet-Res50, PSPNet-Res50, DeepLabv3-Res50 and DeepLabv3-Res101 are tested as target models. The low mIoU of target models means the high transferability of the AEs. Note that we only show the mIoU on the target models since the one on the source model is very low. 

\begin{figure}[t]    
    \centering
    \includegraphics[scale=0.68]{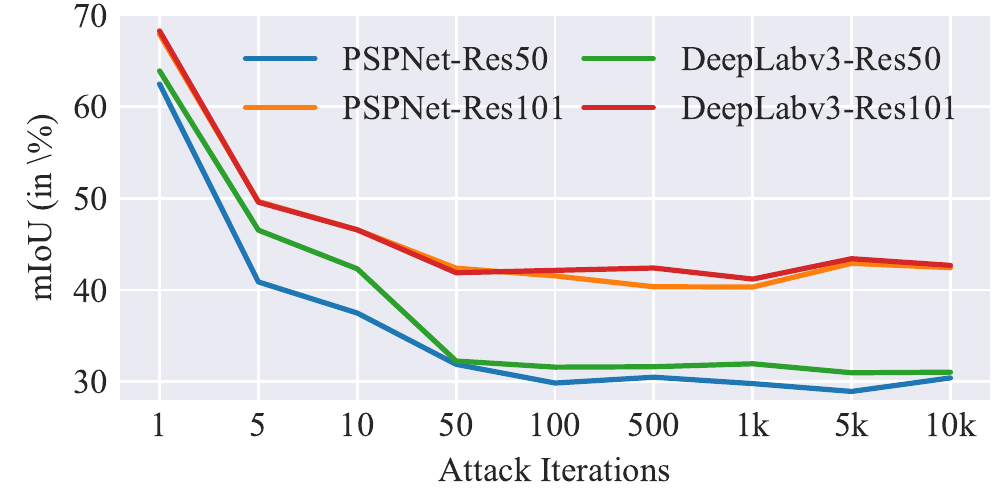}
    \caption{Transferability of adversarial examples on FCN8s-VGG16. No overfitting phenomenon is observed on AEs created on FCN8s-VGG16 even when 10k attack iterations are applied.}
    \label{fig:fcn_10k_overfit}
\end{figure}

When only a few attack iterations are applied, the AEs underfit the source segmentation model. The transferability of the AEs is low, given the high mIoU on target segmentation models in Fig.~\ref{fig:seg_overfit}. In contrast to the observations on classifications, we find that the AEs on FCN8s with VGG backbone do not overfit even when a large number of attack iterations is applied, e.g., 1k iterations. The finding is still true even when 10k attack iterations is applied, as shown in Fig.~\ref{fig:fcn_10k_overfit}. The transferability of AEs on FCN8s-VGG16 can be improved by applying a large number of attack iterations since there is no overfitting observed. The transferability of AEs at the 1k-th iteration is much higher than that reported in~\cite{xie2019improving}. AEs on segmentation models are not difficult to transfer, in contrast to the claims in current literature~\cite{xie2019improving,arnab2018robustness}.

\begin{figure}[t]
    \centering
    \includegraphics[scale=0.68]{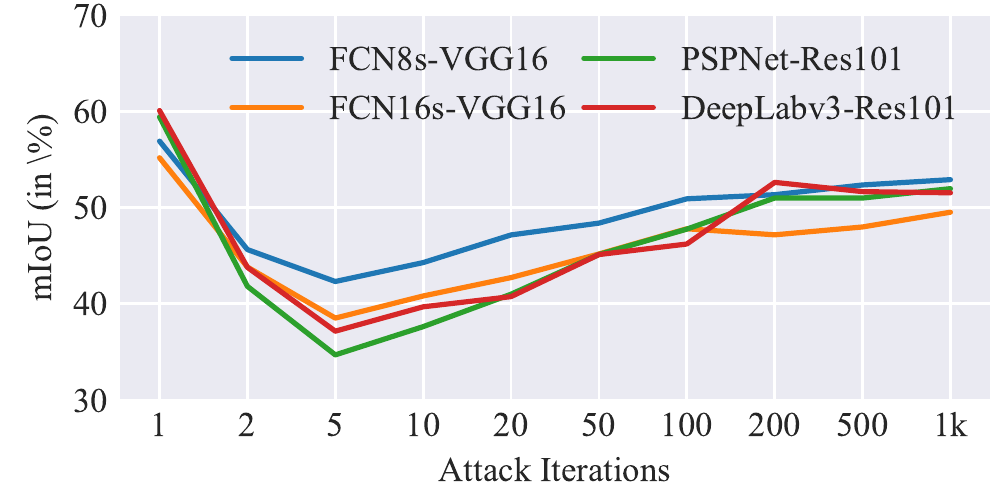}
    \caption{The adversarial examples on DeepLabV3-Res50 also quickly overfit the source segmentation models since the ResNet backbone is also used in this segmentation model.}
    \label{fig:deeplab_overfit}
\end{figure}

In contrast, the AEs on PSPNet-Res50 overfit drastically and quickly. It starts to overfit to source model after only around 5 attack iterations. The drastic overfitting can be attributed to the ResNet backbone, as we analyzed in the last section. Similar drastic overfitting can be observed on PSPNet-Res101 in Fig.~\ref{fig:seg_psp101_overfit} and DeepLabv3-Res50 in Fig.~\ref{fig:deeplab_overfit} since they all are built on ResNet backbone. We assume that the high speed to overfit is caused by the loss used to create AEs, namely, the Equation~\ref{equ:seg_loss}. Specifically, the gradients averaged over all pixels are more accurate to the source model, which make the AEs quickly overfit to it. 

Besides, We compare adversarial perturbations created on classification and segmentation. The perturbations created on classification are often random to the human visual system. However, the ones on segmentation form some visual patterns, although it does not resemble any class (see Appendix). We speculate patterns have to be formed to fool as many pixel classifications as possible in segmentation.

\begin{table*}[t]
\begin{center}
\footnotesize
\setlength\tabcolsep{0.08cm}
\begin{tabular}{c | c | c | ccc cccc}
\toprule
\multicolumn{3}{c|}{Target Models} & FCN8s-VGG16 & FCN16s-VGG16 &  PSPNet-Res50 &  PSPNet-Res101 & DeepLabV3-Res50 & DeepLabV3-Res101\\
\midrule
\multicolumn{3}{c|}{Clean Images} & 69.10 & 66.78 &  78.55 &  79.11 & 78.17 & 80.55\\
\midrule
\multirow{6}{*}{\pbox{20cm}{Source\\Models}} & \multirow{2}{*}{FCN8s-VGG16}  & { FS }  & \underline{0.69} & \underline{0.70}  & 29.79 & 40.32  & 31.95 & 41.19 \\
  & & \textbf{DS} & \textbf{\underline{0.41}} & \textbf{\underline{0.42}} & \textbf{12.21} & \textbf{20.38} & \textbf{9.69} & \textbf{23.70}   \\
\cmidrule{2-9}
  & \multirow{2}{*}{PSPNet-Res50} & FS & 44.17 & 40.26 & \underline{5.09} & 37.06 & \underline{6.57} & 38.98 \\
  & & \textbf{DS} & \textbf{19.95} & \textbf{17.06} & \textbf{\underline{2.07}} & \textbf{16.10} & \textbf{\underline{2.56}}  & \textbf{18.57}  \\
\cmidrule{2-9}
  & \multirow{2}{*}{DeepLabV3-Res50} & FS & 26.19 & 24.34 & \underline{3.63} & 22.01 & \underline{3.14} & 22.58  \\
 & & \textbf{DS} & \textbf{19.63} & \textbf{15.72}  & \textbf{\underline{2.31}} & \textbf{12.32} & \textbf{\underline{2.15}}  & \textbf{13.64}  \\
\bottomrule
\end{tabular}
\end{center}
\caption{Transferring adversarial examples on source segmentation models to target models on VOC. The mIoU of target models on both clean images and adversarial images is shown. For each source model, two methods are applied to create AEs, Fixed-Scale (FS) and our Dynamic-Scale (DS) attack. The mIoU on AEs created with DS is lower, which means that the created AEs are easier to transfer. The mIoU of target models is marked with underlines when they have the same type of backbone architecture as the source model.}
\label{tab:seg_base_voc}
\end{table*}

\begin{table*}[t]
\begin{center}
\footnotesize
\setlength\tabcolsep{0.08cm}
\begin{tabular}{c | c | c | ccc cccc}
\toprule
\multicolumn{3}{c|}{Target Models} & FCN8s-VGG16 & FCN16s-VGG16 &  PSPNet-Res50 &  PSPNet-Res101 & DeepLabV3-Res50 & DeepLabV3-Res101\\
\midrule
\multicolumn{3}{c|}{Clean Images} & 59.3 & 56.86 &  64.9 &  63.01 & 63.5 & 62.8 \\
\midrule
\multirow{6}{*}{\pbox{20cm}{Source\\Models}} & \multirow{2}{*}{FCN8s-VGG16}  & { FS } & \underline{0.26} & \underline{0.27} &  17.95 & 22.39 & 21.83 & 22.60 \\
 & & \textbf{DS} & \textbf{\underline{0.22}} & \textbf{\underline{0.23}} & \textbf{3.72} & \textbf{7.41} & \textbf{3.83} & \textbf{7.75} \\
\cmidrule{2-9}
  & \multirow{2}{*}{PSPNet-Res50} & FS & 24.10 & 22.17 & \underline{3.43} & 24.18 & \underline{5.05} & 25.74 \\
 & & \textbf{DS} & \textbf{10.49} & \textbf{9.23} & \textbf{\underline{0.82}} & \textbf{8.04} & \textbf{\underline{1.36}} & \textbf{9.00} \\
\cmidrule{2-9}
  & \multirow{2}{*}{DeepLabV3-Res50} & FS & 21.78 & 20.23 & \underline{4.76} & 21.72 & \underline{3.92} & 22.23 \\
 & & \textbf{DS} & \textbf{10.61} & \textbf{9.48} & \textbf{\underline{1.23}} & \textbf{7.97} & \textbf{\underline{0.61}} & \textbf{7.11} \\
\bottomrule
\end{tabular}
\end{center}
\caption{Transferring adversarial examples on source segmentation models to target models on Cityscapes. The mIoU on both clean images and adversarial ones is shown. Similar to the conclusion in Table~\ref{tab:seg_base_voc}, the AEs created with DS method achieve higher transferability.}
\label{tab:seg_base_cityscapes}
\end{table*}

\section{Improving Transferability of AEs on Segmentation Models}
In the last section, we find that the AEs created on the segmentation model with VGG backbone do not overfit the source model. The transferability of AEs can be significantly improved by applying a large number of attack iterations, see Fig.~\ref{fig:seg_fcn_overfit} and~\ref{fig:seg_fcn16s_overfit}. However, the transferability is still limited, even when no overfitting is observed on the number of attack iterations.

In semantic segmentation, the segmentation results depend on the size of the input image. To recognize the objects of different scales, dilated convolution and specialized pooling are often applied~\cite{chen2017deeplab,yu2015multi,chen2017deeplab,zhao2017pyramid}. Different models segment objects at different scales. The segmentation models are often built on CNN backbones, which are known to be not invariant to scale~\cite{fawzi2015manitest,henriques2017warped}. Due to the fact, the adversarial perturbation created on a segmentation model may not able to fool other models.

Based on the architectural trait of segmentation models, we propose a simple and effective method to improve the transferability of AEs on Segmentation, where we dynamically scale the input images and ground-truth masks during the attack (dubbed as DS). Concretely, we apply a large number of attack iterations and random resize to the input and ground truth in each attack iteration. By doing this, the created adversarial perturbations can fool segmentation at different scales. In our proposed method, Equation \ref{equ:bim_seg} is applied where we replace the loss term therein with

{\footnotesize
\begin{equation}
L = \frac{1}{{H}\times{W}} \sum_{i=1}^{{H}\times{W}} CE(f_{seg}(\psi_{DS}(\boldsymbol{X}^{adv_{t}}))_i, \psi_{DS}(\boldsymbol{Y}_i)),
\label{equ:seg_baseline}
\end{equation}
}where the $\psi_{DS}(\cdot)$ is a differential operation, which scales both input images and ground-truth masks to different sizes. The scaling ratio in each attack iteration is uniformly sampled from the interval $[1-\lambda, 1+\lambda]$.

A similar approach is also explored to improve the transferability of AEs on classification models. Various image transformations are designed to address the overfitting problem, such as affine transformations~\cite{xie2019improving,dong2019evading,zou2020improving}, and even adversarial transformation~\cite{wu2021improving}. On the one hand, they all assume that the model predicts the same output class when the input is slightly transformed. However, the assumption does not hold on the segmentation models. On the other hand, they all aim to mitigate the overfitting phenomenon on classification models. However, no overfitting is observed on some segmentation models. We aim to overcome the limitation brought by the architectural traits of segmentation models. To this end, we propose a simple, effective scaling transformation for both inputs and ground truth.

\begin{figure*}[t]    
    \begin{subfigure}[b]{\textwidth}
    \centering
    {\footnotesize \hspace{0.4cm} Clean \hspace{0.8cm} FCN8s-VGG16  \hspace{0.4cm} FCN8s-VGG16  \hspace{0.3cm} PSPNet-Res50  \hspace{0.3cm} PSPNet-Res101  \hspace{0.0cm} DeepLabv3-Res50  \hspace{0.0cm} DeepLabv3-Res101  \hspace{0.1cm} Ground-Truth}
    \includegraphics[width=\textwidth]{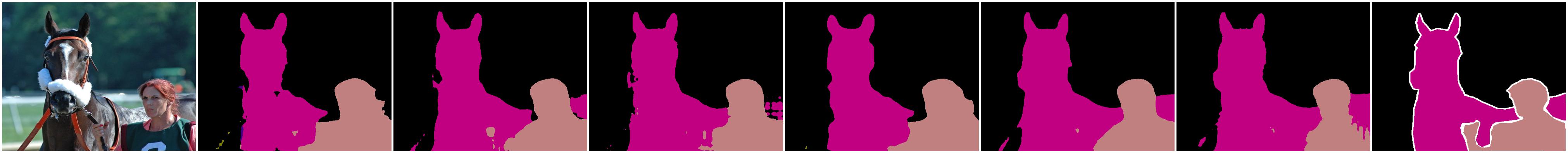}
    \caption{\footnotesize Model Predictions on Clean Images} \vspace{0.6em}
    \label{fig:seg_vis_clean}
    \end{subfigure}  
    \begin{subfigure}[b]{\textwidth}
    \centering
    {\footnotesize \hspace{0.2cm} Adversarial \hspace{0.4cm} FCN8s-VGG16  \hspace{0.4cm} FCN8s-VGG16  \hspace{0.3cm} PSPNet-Res50  \hspace{0.3cm} PSPNet-Res101  \hspace{0.1cm} DeepLabv3-Res50  \hspace{0.0cm} DeepLabv3-Res101  \hspace{0.1cm} Ground-Truth}
    \includegraphics[width=\textwidth]{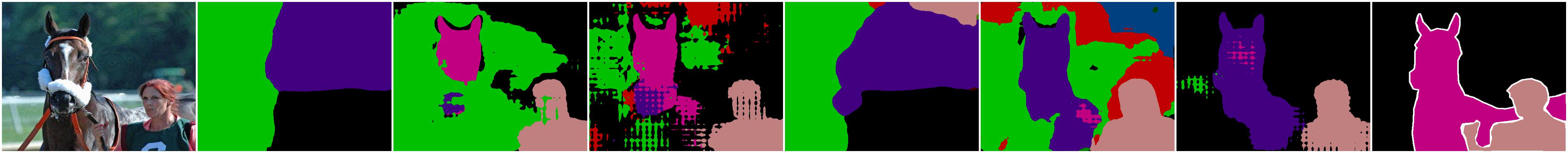}
    \caption{\footnotesize Model Predictions on Adversarial Images Created on FCN8s-VGG with Fixed-Scale Attack} \vspace{0.6em}
    \label{fig:seg_vis_ss}
    \end{subfigure}  
    \begin{subfigure}[b]{\textwidth}
    \centering
    {\footnotesize \hspace{0.2cm} Adversarial \hspace{0.4cm} FCN8s-VGG16  \hspace{0.4cm} FCN8s-VGG16  \hspace{0.3cm} PSPNet-Res50  \hspace{0.3cm} PSPNet-Res101  \hspace{0.1cm} DeepLabv3-Res50  \hspace{0.0cm} DeepLabv3-Res101  \hspace{0.1cm} Ground-Truth}
    \includegraphics[width=\textwidth]{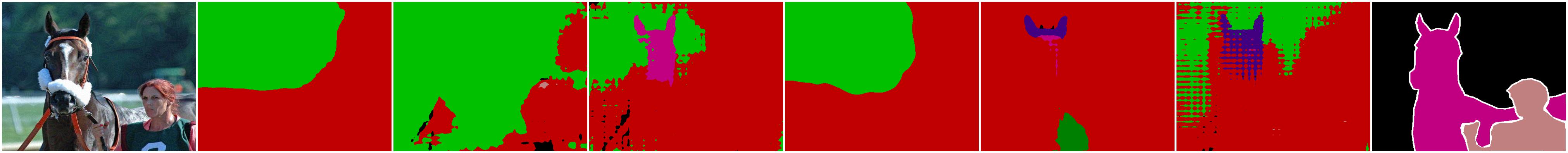} 
    \caption{\footnotesize Model Predictions on Adversarial Images Created on FCN8s-VGG with Dynamic-Scale Attack}
    \label{fig:seg_vis_ds}
    \end{subfigure} \vspace{-0.25cm}
    \caption{Visualization of Clean Images, Adversarial Images, Segmentation Predictions, and Ground Truth. FCN8s-VGG is taken as the source model. The AEs created with Dynamic-Scale attack transfer better to other segmentation models.  More figures are in Appendix.}
    \label{fig:seg_vis_adv}
\end{figure*}

\paragraph{Experimental results and analysis.} We use the same experimental setting as in the last section. Concretely, we apply FCN8s-VGG16, PSPNet-Res50, and DeepLabv3-Res50 as source models, respectively. The segmentation models with different backbones are applied as target segmentation models. The mIoU scores of target models on clean images and adversarial images are reported in Table~\ref{tab:seg_base_voc}.

When the size of the input image is fixed during the attack, i.e. a single fixed scale attack (FS), we report the best mIoU of FCNet with VGG backbone at 1k iteration and the best mIoU of other models with ResNet backbone at the 5-th attack iteration, as shown in Fig.~\ref{fig:seg_overfit}. When our dynamic scaling (DS) is applied during the attack, the AEs on ResNet do not overfit the source model either since the input image is scaled to a different size in each attack iteration. In the case of our DS, we report the mIoU of all target segmentation models at 1k-th iteration. The target models show much less mIoU on the adversarial images, which means the AEs are highly transferable. E.g., the mIoU of DeepLabv3-Res50 can be reduced to 9.69. Our method also works well on Cityscapes dataset, shown in Table~\ref{tab:seg_base_cityscapes}. The high transferability achieved by our simple method reveal that adversarial example on segmentation models can be easy to transfer.

The clean images, adversarial images, and the predictions made by target segmentation models on them are visualized in Fig.~\ref{fig:seg_vis_adv}. In Fig.~\ref{fig:seg_vis_clean}, we show the predictions made by different segmentation models on the clean images. The predictions are close to the ground truth. When attacking with FS on FCN8s-VGG16, the predictions on the source model can be misled by the imperceptible adversarial perturbations, as shown in Fig.~\ref{fig:seg_vis_ss}. The AEs can also partly fool other segmentation models. When DS is applied during the attack, the created adversarial perturbation transfer better to other segmentation models, as shown in Fig.~\ref{fig:seg_vis_ds}.

\begin{table}[t]
\begin{center}
\footnotesize
\setlength\tabcolsep{0.14cm}
\begin{tabular}{c | c | c | cc}
\toprule
\multicolumn{3}{c|}{\multirow{2}{*}{Target Models}} &  \multicolumn{2}{c}{Adversarial Training} \\
\cmidrule{4-5}
\multicolumn{3}{c|}{}  &  PSPNet-Res50 &  DeepLabV3-Res50 \\
\midrule
\multicolumn{3}{c|}{Clean Image} & 73.46 & 71.42 \\
\midrule
\multirow{4}{*}{\pbox{20cm}{Source\\Models}} &\multirow{2}{*}{FCN8s-VGG16}  & FS  & 67.66 & 60.67 \\
 & & \textbf{DS} & \textbf{66.89} & \textbf{58.99} \\
\cmidrule{2-5}
  & \multirow{2}{*}{PSPNet-Res50} & FS & 65.70 & 59.48 \\
  & & \textbf{DS} & \textbf{64.41} & \textbf{57.58} \\
\bottomrule
\end{tabular} \vspace{-0.2cm}
\end{center}
\caption{Transferability of adversarial examples created on adversarially trained segmentation models. The adversarial images created with DS attack transfer better than the ones with FS attack.} \vspace{-0.2cm}
\label{tab:seg_at_voc}
\end{table}

\begin{figure*}[t]    
    \begin{subfigure}[b]{0.48\textwidth}
    \centering
    \includegraphics[scale=0.65]{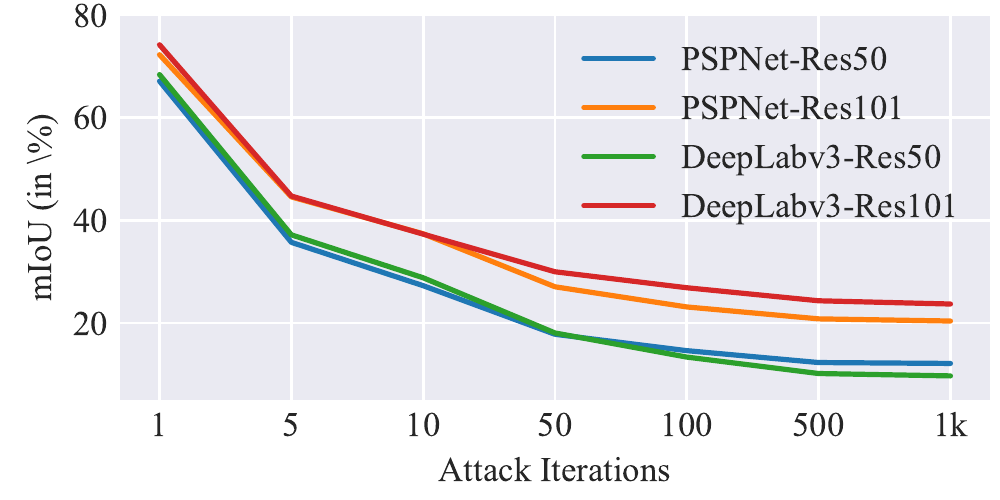}
    \caption{\footnotesize FCN8s-VGG16 as Source Model}
    \end{subfigure}  \hspace{1.3em}
    \begin{subfigure}[b]{0.48\textwidth}
    \centering
    \includegraphics[scale=0.65]{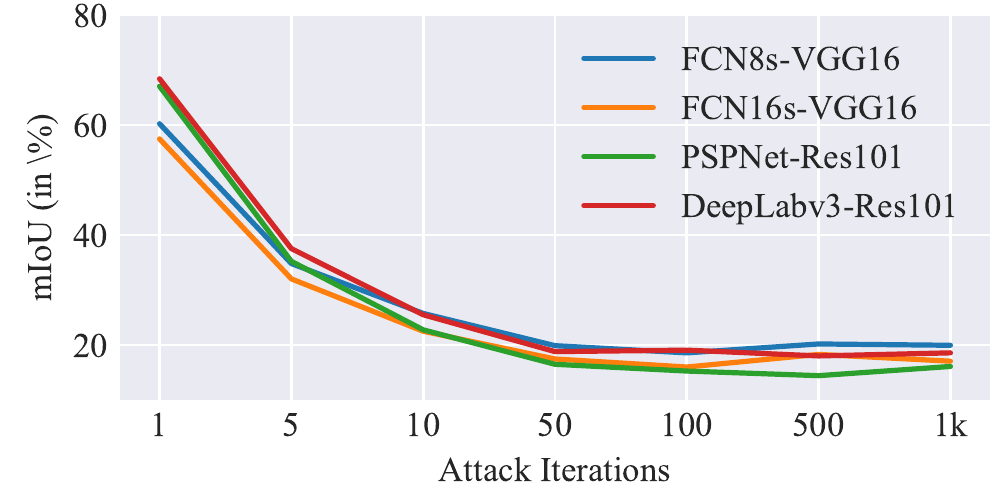}
    \caption{\footnotesize PSPNet-Res50 as Source Model}
    \end{subfigure}
    \caption{Transferability of adversarial examples created with Dynamic-Scale attack method at different attack iterations. When Dynamic-Scale attack is applied, no overfitting is observed on AEs even on the source model with ResNet backbone.}
    \label{fig:iter_study}
\end{figure*}

\vspace{0.15cm}\noindent\textbf{Transferability of AEs to adversarially trained segmentation models. } One of the most effective defense methods in classification is adversarial training~\cite{Biggio2013EvasionAA,szegedy2013intriguing,goodfellow2014explaining,madry2018towards}, where the AEs created during training are injected into the training data to train the models. The standard adversarial training method can be generalized to segmentation~\cite{xu2021dynamic}. We also test our method on the adversarially trained models, shown in Table~\ref{tab:seg_at_voc}. In general, the robustness of segmentation models is improved significantly by adversarial training. Our AEs with DS is still more effective to fool the adversarially trained models than the ones with FS. This study also indicates the necessity of an adversarial defense method to defend against transfer-based attacks in segmentation.

\vspace{0.1cm}\noindent\textbf{Transferability of AEs to different scales. } Previous work~\cite{arnab2018robustness} shows that the AEs created on a scale do not transfer well to segmentation on a different scale on the same model. As shown in Table~\ref{tab:seg_scale_voc}, the AEs crated on FCN8s-VGG16 under FS attack on the scale of 100\% are much less effective to fool the segmentation on the scale of 50\%. Our method improves the effectiveness of the created AEs significantly even when they are segmented from a different scale. The conclusion is also true on the PSPNet-Res50.
\begin{table}[!h]
\begin{center}
\footnotesize
\setlength\tabcolsep{0.18cm}
\begin{tabular}{c | c | p{0.9cm} p{0.9cm} p{0.8cm} c}
\toprule
  \multicolumn{2}{c|}{} &  50\% &  75\% &  100\%  & Multi-Scale  \\
\midrule
 \multirow{2}{*}{FCN8s-VGG16}  & FS  & 50.74 & 10.67 & 0.97 & 1.07 \\
 & \textbf{DS} & \textbf{10.40} & \textbf{0.87} &  \textbf{0.40}  &  \textbf{0.52} \\
\midrule
  \multirow{2}{*}{PSPNet-Res50} & FS  & 64.0 & 56.11 & 2.17  & 6.52 \\
 & \textbf{DS} & \textbf{21.85} & \textbf{2.91} & \textbf{2.07} & \textbf{2.72} \\
\bottomrule
\end{tabular} \vspace{-0.2cm}
\end{center}
\caption{Transferring adversarial examples to multi-scale segmentation. The AEs created with DS attack transfer well to segmentation at different scales as well as multi-scale segmentation.}
\label{tab:seg_scale_voc}
\end{table}

\vspace{0.1cm}\noindent\textbf{Ablation on the number of attack iterations. } We report the accuracy of target classifiers on the AEs created in different attack iterations in Fig.~\ref{fig:iter_study}. When our DS is applied, no overfitting phenomenon is observed, even on segmentation models with ResNet backbone. The more attack iterations are applied, the better the transferability is. The observation is still true when 10k attack iterations are applied.

Note that, in the research community, currently, the efficiency to create transferable AEs is less concerned compared to the effectiveness. In fact, our approach has already achieved good transferability in the $200$-th attack step. We speculate that the efficiency of our approach can be improved by adapting the step size during the attack. We leave the further exploration in future work.

\begin{figure}[t]    
    \centering
    \includegraphics[scale=0.72]{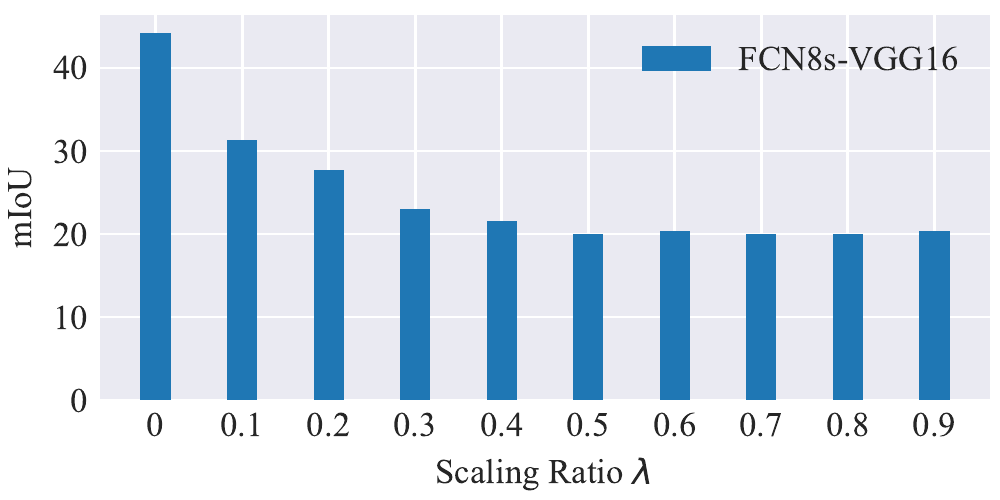}
    \caption{Ablation study of scaling ratio of DS attack method. An increase of the scaling ratio leads to a low mIoU of target models, i.e., better transferability of the AEs. When it is bigger than 0.5, the transferability is not sensitive to the scaling ratio.}
    \label{fig:ratio_study}
\end{figure}

\vspace{0.1cm}\noindent\textbf{Ablation on scaling ratios. } In our method, a scaling operation is applied. We also study the sensitivity of the transferability of AEs to the scaling ratio. As shown in Fig.~\ref{fig:ratio_study}, we report the mIoU of FCN8s-VGG on the AEs created on PSPNet-Res50. We can observe that when the scaling ratio is set to 0.1, the transferability of adversarial examples can be improved by a large margin, compared to the score with no scaling operation ($\lambda$=0). The transferability can be further improved by applying a bigger scale ratio until 0.5. When a bigger scale ratio is applied during the attack, the transferability is not sensitive to the scaling ratio anymore. The ratio of 0.5 is also used across this paper.

\section{Conclusion}
This work focuses on the transferability of AEs on segmentation. We first investigate the transferability of AEs created on classifiers that correspond to segmentation backbones. We find that all classification models suffer from overfitting of AEs and the degree of overfitting depends on the model architectures. Our further investigation on AEs on segmentation shows the AEs do not always overfit to source model. When non-overfitting is presented, the AEs can be made more transferable by applying a large number of attack iterations. Furthermore, based on our investigation and architectural traits of segmentation models, we propose a simple, yet effective method, which improves the transferability significantly. Our work is the first to show that AEs on segmentation models can be easy to transfer. Due to the feasible transfer-based black-box attacks on them, segmentation models also face potential black-box attack threats in safety-critical applications.

\newpage
{
\bibliographystyle{ieee_fullname}
\bibliography{egbib}

\begin{thebibliography}{10}\itemsep=-1pt

\bibitem{arnab2018robustness}
Anurag Arnab, Ondrej Miksik, and Philip~HS Torr.
\newblock On the robustness of semantic segmentation models to adversarial
  attacks.
\newblock In {\em CVPR}, 2018.

\bibitem{Biggio2013EvasionAA}
Battista Biggio, Igino Corona, Davide Maiorca, Blaine Nelson, Nedim Srndic,
  Pavel Laskov, Giorgio Giacinto, and Fabio Roli.
\newblock Evasion attacks against machine learning at test time.
\newblock In {\em arXiv:1708.06131}, 2013.

\bibitem{chen2017deeplab}
Liang-Chieh Chen, George Papandreou, Iasonas Kokkinos, Kevin Murphy, and Alan~L
  Yuille.
\newblock Deeplab: Semantic image segmentation with deep convolutional nets,
  atrous convolution, and fully connected crfs.
\newblock {\em IEEE transactions on pattern analysis and machine intelligence
  (TPAMI)}, 2017.

\bibitem{chen2017rethinking}
Liang-Chieh Chen, George Papandreou, Florian Schroff, and Hartwig Adam.
\newblock Rethinking atrous convolution for semantic image segmentation.
\newblock In {\em arXiv:1706.05587}, 2017.

\bibitem{cordts2016cityscapes}
Marius Cordts, Mohamed Omran, Sebastian Ramos, Timo Rehfeld, Markus Enzweiler,
  Rodrigo Benenson, Uwe Franke, Stefan Roth, and Bernt Schiele.
\newblock The cityscapes dataset for semantic urban scene understanding.
\newblock In {\em CVPR}, 2016.

\bibitem{deng2009imagenet}
Jia Deng, Wei Dong, Richard Socher, Li-Jia Li, Kai Li, and Li Fei-Fei.
\newblock Imagenet: A large-scale hierarchical image database.
\newblock In {\em CVPR}, 2009.

\bibitem{dong2018boosting}
Yinpeng Dong, Fangzhou Liao, Tianyu Pang, Hang Su, Jun Zhu, Xiaolin Hu, and
  Jianguo Li.
\newblock Boosting adversarial attacks with momentum.
\newblock In {\em CVPR}, 2018.

\bibitem{dong2019evading}
Yinpeng Dong, Tianyu Pang, Hang Su, and Jun Zhu.
\newblock Evading defenses to transferable adversarial examples by
  translation-invariant attacks.
\newblock In {\em CVPR}, 2019.

\bibitem{everingham2010pascal}
Mark Everingham, Luc Van~Gool, Christopher~KI Williams, John Winn, and Andrew
  Zisserman.
\newblock The pascal visual object classes (voc) challenge.
\newblock {\em International journal of computer vision (IJCV)}, 2010.

\bibitem{fawzi2015manitest}
Alhussein Fawzi and Pascal Frossard.
\newblock Manitest: Are classifiers really invariant?
\newblock In {\em arXiv:1507.06535}, 2015.

\bibitem{fischer2021scalable}
Marc Fischer, Maximilian Baader, and Martin Vechev.
\newblock Scalable certified segmentation via randomized smoothing.
\newblock In {\em ICML}, 2021.

\bibitem{goodfellow2014explaining}
Ian~J Goodfellow, Jonathon Shlens, and Christian Szegedy.
\newblock Explaining and harnessing adversarial examples.
\newblock In {\em ICLR}, 2015.

\bibitem{guo2020backpropagating}
Yiwen Guo, Qizhang Li, and Hao Chen.
\newblock Backpropagating linearly improves transferability of adversarial
  examples.
\newblock In {\em NeurIPS}, 2020.

\bibitem{he2016deep}
Kaiming He, Xiangyu Zhang, Shaoqing Ren, and Jian Sun.
\newblock Deep residual learning for image recognition.
\newblock In {\em CVPR}, 2016.

\bibitem{he2019non}
Xiang He, Sibei Yang, Guanbin Li, Haofeng Li, Huiyou Chang, and Yizhou Yu.
\newblock Non-local context encoder: Robust biomedical image segmentation
  against adversarial attacks.
\newblock In {\em AAAI}, 2019.

\bibitem{hendrik2017universal}
Jan Hendrik~Metzen, Mummadi Chaithanya~Kumar, Thomas Brox, and Volker Fischer.
\newblock Universal adversarial perturbations against semantic image
  segmentation.
\newblock In {\em ICCV}, 2017.

\bibitem{henriques2017warped}
Joao~F Henriques and Andrea Vedaldi.
\newblock Warped convolutions: Efficient invariance to spatial transformations.
\newblock In {\em ICML}, 2017.

\bibitem{howard2017mobilenets}
Andrew~G Howard, Menglong Zhu, Bo Chen, Dmitry Kalenichenko, Weijun Wang,
  Tobias Weyand, Marco Andreetto, and Hartwig Adam.
\newblock Mobilenets: Efficient convolutional neural networks for mobile vision
  applications.
\newblock In {\em arXiv:1704.04861}, 2017.

\bibitem{huang2019enhancing}
Qian Huang, Isay Katsman, Horace He, Zeqi Gu, Serge Belongie, and Ser-Nam Lim.
\newblock Enhancing adversarial example transferability with an intermediate
  level attack.
\newblock In {\em ICCV}, 2019.

\bibitem{inkawhich2020perturbing}
Nathan Inkawhich, Kevin~J Liang, Binghui Wang, Matthew Inkawhich, Lawrence
  Carin, and Yiran Chen.
\newblock Perturbing across the feature hierarchy to improve standard and
  strict blackbox attack transferability.
\newblock In {\em NeurIPS}, 2020.

\bibitem{kaymak2019brief}
{\c{C}}a{\u{g}}r{\i} Kaymak and Ay{\c{s}}eg{\"u}l U{\c{c}}ar.
\newblock A brief survey and an application of semantic image segmentation for
  autonomous driving.
\newblock In {\em Handbook of Deep Learning Applications}. 2019.

\bibitem{kurakin2016adversarial}
Alexey Kurakin, Ian Goodfellow, Samy Bengio, et~al.
\newblock Adversarial examples in the physical world.
\newblock In {\em ICLR}, 2016.

\bibitem{li2020learning}
Yingwei Li, Song Bai, Yuyin Zhou, Cihang Xie, Zhishuai Zhang, and Alan Yuille.
\newblock Learning transferable adversarial examples via ghost networks.
\newblock In {\em AAAI}, 2020.

\bibitem{lin2019nesterov}
Jiadong Lin, Chuanbiao Song, Kun He, Liwei Wang, and John~E Hopcroft.
\newblock Nesterov accelerated gradient and scale invariance for adversarial
  attacks.
\newblock In {\em ICLR}, 2020.

\bibitem{long2015fully}
Jonathan Long, Evan Shelhamer, and Trevor Darrell.
\newblock Fully convolutional networks for semantic segmentation.
\newblock In {\em CVPR}, 2015.

\bibitem{madry2018towards}
Aleksander Madry, Aleksandar Makelov, Ludwig Schmidt, Dimitris Tsipras, and
  Adrian Vladu.
\newblock Towards deep learning models resistant to adversarial attacks.
\newblock In {\em ICLR}, 2018.

\bibitem{milletari2016v}
Fausto Milletari, Nassir Navab, and Seyed-Ahmad Ahmadi.
\newblock V-net: Fully convolutional neural networks for volumetric medical
  image segmentation.
\newblock In {\em 3DV}, 2016.

\bibitem{Nesterov1983AMF}
Yurii Nesterov.
\newblock A method for unconstrained convex minimization problem with the rate
  of convergence o(1/k caret 2).
\newblock {\em Doklady an ussr}, 1983.

\bibitem{nesti2021evaluating}
Federico Nesti, Giulio Rossolini, Saasha Nair, Alessandro Biondi, and Giorgio
  Buttazzo.
\newblock Evaluating the robustness of semantic segmentation for autonomous
  driving against real-world adversarial patch attacks.
\newblock In {\em arXiv:2108.06179}, 2021.

\bibitem{paschali2018generalizability}
Magdalini Paschali, Sailesh Conjeti, Fernando Navarro, and Nassir Navab.
\newblock Generalizability vs. robustness: investigating medical imaging
  networks using adversarial examples.
\newblock In {\em MICCAI}, 2018.

\bibitem{NEURIPS2019_9015}
Adam Paszke, Sam Gross, Francisco Massa, Adam Lerer, James Bradbury, Gregory
  Chanan, Trevor Killeen, Zeming Lin, Natalia Gimelshein, Luca Antiga, Alban
  Desmaison, Andreas Kopf, Edward Yang, Zachary DeVito, Martin Raison, Alykhan
  Tejani, Sasank Chilamkurthy, Benoit Steiner, Lu Fang, Junjie Bai, and Soumith
  Chintala.
\newblock Pytorch: An imperative style, high-performance deep learning library.
\newblock In {\em NeurIPS}, 2019.

\bibitem{polyak1964some}
Boris~T Polyak.
\newblock Some methods of speeding up the convergence of iteration methods.
\newblock {\em ussr computational mathematics and mathematical physics}, 1964.

\bibitem{simonyan2014very}
Karen Simonyan and Andrew Zisserman.
\newblock Very deep convolutional networks for large-scale image recognition.
\newblock In {\em ICLR}, 2015.

\bibitem{GoogleNetSzegedyLJSRAEVR14}
Christian Szegedy, Wei Liu, Yangqing Jia, Pierre Sermanet, Scott Reed, Dragomir
  Anguelov, Dumitru Erhan, Vincent Vanhoucke, and Andrew Rabinovich.
\newblock Going deeper with convolutions.
\newblock In {\em CVPR}, 2015.

\bibitem{szegedy2013intriguing}
Christian Szegedy, Wojciech Zaremba, Ilya Sutskever, Joan Bruna, Dumitru Erhan,
  Ian Goodfellow, and Rob Fergus.
\newblock Intriguing properties of neural networks.
\newblock In {\em ICLR}, 2014.

\bibitem{tan2019mnasnet}
Mingxing Tan, Bo Chen, Ruoming Pang, Vijay Vasudevan, Mark Sandler, Andrew
  Howard, and Quoc~V Le.
\newblock Mnasnet: Platform-aware neural architecture search for mobile.
\newblock In {\em CVPR}, 2019.

\bibitem{wang2020unified}
Xin Wang, Jie Ren, Shuyun Lin, Xiangming Zhu, Yisen Wang, and Quanshi Zhang.
\newblock A unified approach to interpreting and boosting adversarial
  transferability.
\newblock In {\em ICLR}, 2021.

\bibitem{wu2020skip}
Dongxian Wu, Yisen Wang, Shu-Tao Xia, James Bailey, and Xingjun Ma.
\newblock Skip connections matter: On the transferability of adversarial
  examples generated with resnets.
\newblock In {\em ICLR}, 2020.

\bibitem{wu2021improving}
Weibin Wu, Yuxin Su, Michael~R Lyu, and Irwin King.
\newblock Improving the transferability of adversarial samples with adversarial
  transformations.
\newblock In {\em CVPR}, 2021.

\bibitem{xiao2018characterizing}
Chaowei Xiao, Ruizhi Deng, Bo Li, Fisher Yu, Mingyan Liu, and Dawn Song.
\newblock Characterizing adversarial examples based on spatial consistency
  information for semantic segmentation.
\newblock In {\em ECCV}, 2018.

\bibitem{xie2017adversarial}
Cihang Xie, Jianyu Wang, Zhishuai Zhang, Yuyin Zhou, Lingxi Xie, and Alan
  Yuille.
\newblock Adversarial examples for semantic segmentation and object detection.
\newblock In {\em ICCV}, 2017.

\bibitem{xie2019improving}
Cihang Xie, Zhishuai Zhang, Yuyin Zhou, Song Bai, Jianyu Wang, Zhou Ren, and
  Alan~L Yuille.
\newblock Improving transferability of adversarial examples with input
  diversity.
\newblock In {\em CVPR}, 2019.

\bibitem{xu2021dynamic}
Xiaogang Xu, Hengshuang Zhao, and Jiaya Jia.
\newblock Dynamic divide-and-conquer adversarial training for robust semantic
  segmentation.
\newblock In {\em ICCV}, 2021.

\bibitem{yu2015multi}
Fisher Yu and Vladlen Koltun.
\newblock Multi-scale context aggregation by dilated convolutions.
\newblock In {\em ICLR}, 2016.

\bibitem{zhao2017pyramid}
Hengshuang Zhao, Jianping Shi, Xiaojuan Qi, Xiaogang Wang, and Jiaya Jia.
\newblock Pyramid scene parsing network.
\newblock In {\em CVPR}, 2017.

\bibitem{zou2020improving}
Junhua Zou, Zhisong Pan, Junyang Qiu, Xin Liu, Ting Rui, and Wei Li.
\newblock Improving the transferability of adversarial examples with
  resized-diverse-inputs, diversity-ensemble and region fitting.
\newblock In {\em ECCV}, 2020.

\end{thebibliography}
}

\newpage
\appendix

\end{document}